\pdfoutput=1

\documentclass[11pt]{article}

\usepackage[]{acl}

\usepackage[]{xcolor}
\usepackage{times}
\usepackage{latexsym}
\usepackage[T1]{fontenc}
\usepackage[utf8]{inputenc}
\usepackage{microtype}
\usepackage[utf8]{inputenc} 
\usepackage[T1]{fontenc}    
\usepackage{hyperref}       
\usepackage{url}            
\usepackage{booktabs}       
\usepackage{amsfonts}       
\usepackage{nicefrac}       
\usepackage{microtype}      
\usepackage{changepage}
\usepackage{xargs}          
\usepackage{wrapfig,lipsum,booktabs}
\usepackage{enumitem}
\usepackage{longtable}
\usepackage{makecell}
\usepackage{graphicx}
\usepackage{subcaption}

\usepackage{changepage}
\usepackage[scaled=1]{couriers}

\title{A Recipe for Arbitrary Text Style Transfer with Large Language Models}

\author{
Emily Reif\textsuperscript{1}\thanks{\ \ Equal contribution}\ \ \ \
Daphne Ippolito\textsuperscript{1,2}\textsuperscript{*}\ \ \ 
Ann Yuan\textsuperscript{1}\ \ \
Andy Coenen\textsuperscript{1}\\
\textbf{Chris Callison-Burch\textsuperscript{2}\ \ \
Jason Wei\textsuperscript{1}}\\
\textsuperscript{1}Google Research \ \ \
\textsuperscript{2}University of Pennsylvania\\
 \{{\tt ereif, annyuan, andycoenen, jasonwei\}@google.com}\\
 \{{\tt daphnei, ccb\}@seas.upenn.edu}
}

\begin{document}
%

\maketitle

\begin{abstract}

In this paper, we leverage large language models (LMs) to perform zero-shot text style transfer.
We present a prompting method that we call \textit{augmented zero-shot learning}, which frames style transfer as a sentence rewriting task and requires only a natural language instruction, without model fine-tuning or exemplars in the target style.
Augmented zero-shot learning is simple and demonstrates promising results not just on standard style transfer tasks such as sentiment, but also on natural language transformations such as ``make this melodramatic'' or ``insert a metaphor.''

\end{abstract}

\section{Introduction}

Text style transfer is the task of rewriting text to incorporate additional or alternative stylistic elements while preserving the overall semantics and structure.
Although style transfer has garnered increased interest due to the success of deep learning, these approaches usually require a substantial amount of labeled training examples, either as parallel text data \citep{zhu-etal-2010-monolingual,rao-tetreault-2018-dear} or non-parallel text data of a single style. \citep{li-etal-2018-delete,jin-etal-2019-imat, liu2020revision,style-transfer-as-paraphrase-2020}. 
Even bleeding-edge approaches that tackle the challenging problem of label-free style transfer are limited in that they require at least several exemplar sentences that dictate a given target style \citep{DBLP:journals/corr/abs-1905-11975,DBLP:journals/corr/abs-2010-03802}.
Hence, recent survey papers have identified a need for new methods that both reduce the training data requirements and expand the scope of styles supported \citep{DBLP:journals/corr/abs-2011-00416,DBLP:journals/corr/abs-2010-12742}.

\begin{figure}[t]
     \includegraphics[width=0.99\linewidth]{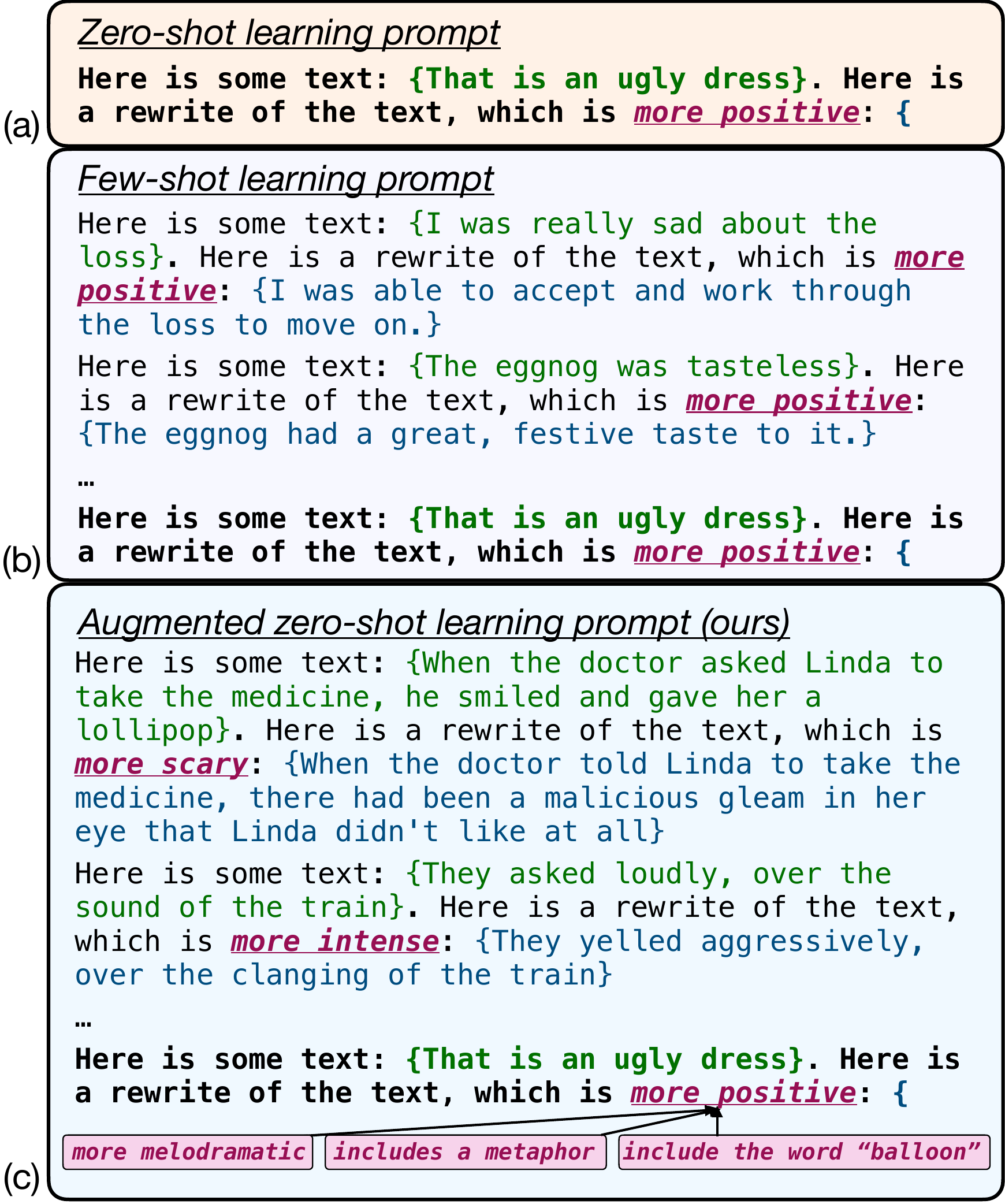}
     \vspace{-1.5mm}
  \caption{Zero-shot, few-shot, and augmented zero-shot prompts for style transfer. The boldface text is the zero-shot prompt, and the plain text is the additional priming sequence. The full prompts used in this paper are shown in Table \ref{tab:fullprompt}. We encourage readers to examine the outputs of our model at \url{https://bit.ly/3fLDuci}.}
  \label{prompts}
\vspace{-3.5mm}
\end{figure}


In this work, we present \textit{augmented zero-shot learning}, a prompting method that allows large language models to perform text style transfer to arbitrary styles, without any exemplars in the target style.
Our method builds on prior work showing that sufficiently large LMs such as GPT-3 can perform various tasks ranging from classification to translation, simply by choosing a clever prompt to prepend to the input text for which the model is asked to continue \citep{DBLP:journals/corr/abs-2005-14165,branwen2020gpt}. 
Using a single prompt that provides several demonstrations of sentences being ``rewritten'' to meet a desired condition, language models can extrapolate and rewrite text in unseen styles.
We are thus able to perform style transfer to arbitrary styles such as ``\textit{make this sentence more comic}'' or ``\textit{include the word balloon}.''

Augmented zero-shot learning is simple and facilitates the application of style transfer to a wider range of styles than existing work.
Our contributions are the following.

\begin{enumerate}[leftmargin=*,noitemsep,nolistsep]
    \item We propose a recipe for style transfer using large LMs that is label-free, training-free, and intuitively controllable.
    \item Via human evaluation, we find that our method achieves strong performance on both standard and non-standard style transfer tasks. We also compare our approach for sentiment transfer with prior methods using automatic evaluation.
    \item We explore real-world desired style transfers generated from users of a text editing UI that implements our method.
\end{enumerate}

\section{Augmented zero-shot prompting}

Although large LMs are trained only for continuation, recent work has shown that they can perform a variety of NLP tasks by expressing the task as a prompt that encourages the model to output the desired answer as the continuation \cite[][\textit{inter alia}; see \citet{,liu2021pre} for a survey]{puri2019zero,weller-etal-2020-learning,DBLP:journals/corr/abs-2005-14165,schick-schutze-2021-just}. 
The simplest approach, \textbf{zero-shot prompting}, directly uses natural language to ask the large LM to perform a task, as shown in Figure \ref{prompts}a. Zero-shot prompting, however, can be prone to failure modes such as not returning well-formatted or logical outputs (see $\S$\ref{section:limitations}).
\textbf{Few-shot prompting}, as shown in Figure \ref{prompts}b, has been shown to achieve higher performance, but requires exemplars for the exact task that we want the model to perform. 
Such few-shot examples can be easily obtained if the desired style transformation is known ahead of time, but this ultimately limits style transfer to a set of pre-specified style tasks.

To remove the need for these labeled exemplars for each style transfer task, we propose \textit{augmented zero-shot learning}, a method for performing multi-task style transfer using a single set of exemplars.
Instead of prompting the model with exemplars specific to the exact style transfer task we wish to perform, we prompt the model with examples of a variety of sentence rewriting operations, as shown in Figure \ref{prompts}c. 
This intuition is inspired by \citet{reynolds2021prompt}'s observation that successful prompts constrain the behavior of the large LM away from failure modes---in our case, we aim to preserve the flexibility of a zero shot prompt while encouraging the model to produce outputs of a specific template.
We keep the the format of the exemplars constant and insert the desired sentence transformation into the same format.
In this way, the augmented zero-shot formulation supports arbitrary sentence rewriting tasks without the need to write any task-specific exemplars.
Thus, it works for a wide range of styles, including modifying the text to be ``\textit{more melodramatic,}'' ``\textit{insert a metaphor,}'' or ``\textit{include the word balloon.}''


\section{Experimental Setup}

\begin{table}[tbp]
\setlength{\tabcolsep}{3.3pt}
  \small
    \label{fig:visualizer_results}
    \begin{tabular}{p{0.25\linewidth}>{\scriptsize}p{0.7\linewidth}}
    \toprule
    Original & \texttt{There, in the middle of the street, stood an old man in a weatherbeaten brown coat.} \\
    \hline
    More\newline descriptive &
    \texttt{The man's brown trenchcoat was coated with snow, and as he leaned against the telephone booth, he cast a long shadow across the road in the late afternoon sun.}\\
    \hline
    Include a\newline metaphor &
    \texttt{There, in the middle of the street, stood an old man whose face resembled the weatherworn bark of a gnarled tree.}\\
    \hline
    Include the \newline word ``\textit{park}'' & \texttt{There, in the middle of Central Park, stood an old man in a weatherbeaten brown coat.} \\
    \hline
    Include the \newline word ``\textit{balloon}'' & \texttt{There, in the middle of the street, stood an old man with several colourful balloons tied to the straps of his coat.} \\
    \hline
    More comic & \texttt{There, in the middle of the street, stood an old man in a raggedy brown coat that was much too big for him.} \\
    \hline
    More melodramatic & \texttt{There, in the middle of the road, stood a grizzled old man, the light of life faded from his sunken eyes.}\\
    \bottomrule
    \end{tabular}%
    \vspace{-1mm}
    \caption{Example style transfer outputs from augmented zero-shot learning for non-standard styles. }  
  \label{tab:style-transfer-examples}%
  \vspace{-2mm}
\end{table}%

\paragraph{Style transfer tasks.}
We consider six style transfer tasks that we deem non-standard, listed in Table \ref{tab:style-transfer-examples}.
These styles were chosen to be representative of most frequent style adjustments made by users of an AI-assisted text editor that employs our method (discussed further in $\S$\ref{subsec:potential}). 
As source sentences, we use 50 sentences randomly drawn from the Reddit Writing Prompts validation set \citep{fan2018hierarchical}, excluding those that already clearly exhibited one of the styles or were ungrammatical/incoherent.
We use human evaluation for these styles, since not all styles have readily available classifiers.

We also evaluate our method on two standard style transfer tasks: sentiment and formality.
We use the Yelp polarity dataset \citep{zhangCharacterlevelConvolutionalNetworks2015} for sentiment and Grammarly's Yahoo Answers Formality Corpus (GYAFC) dataset for formality \citep{rao-tetreault-2018-dear}.\footnote{Hosted by \citet{DBLP:conf/ijcai/LuoLZYCSS19}.}
These datasets allow us to evaluate performance of augmented zero-shot learning in the context of prior supervised methods which have been used on these tasks.

\paragraph{Model.}
Augmented zero-shot learning requires a large language model.
We primarily use LaMDA, a left-to-right decoder-only transformer language model \citep{DBLP:journals/corr/VaswaniSPUJGKP17} with a non-embedding parameter count of 137B \citep{thoppilan2022lamda}. 
The pre-trained LaMDA model, which we refer to as \textit{LLM}, was trained on a corpus comprising 1.95B public web documents, including forum and dialog data and Wikipedia.
The dataset was tokenized into 2.49T BPE tokens with a SentencePiece vocabulary size of 32K \citep{DBLP:journals/corr/abs-1808-06226}.
We also use \textit{LLM-Dialog}, the final LaMDA model which was finetuned on a curated, high-quality subset of data identified to be in a conversational format.
Decoding was done with top-$k$=40.
To show that the success of augmented zero-shot learning is not restricted to these two large LMs, we also perform experiments with GPT-3 (Table \ref{tab:candidate_select}).
For GPT-3, decoding was done with nucleus sampling using $p$=0.6 \citep{holtzman2019curious}.

The prompts used for \textit{LLM} and GPT-3 are shown in Figure \ref{prompts}.
For \textit{LLM-Dialog}, the prompt was instead formulated as a conversation between one agent who is requesting rewrites and another who is performing the rewrites.
See Table \ref{tab:fullprompt} in the Appendix for the full non-abbreviated prompts.

\section{Results}
\subsection{Non-Standard Styles}
For our six non-standard styles, we asked six professional raters to assess  <input sentence, target style, output sentence> tuples. These raters are fluent in English, live in India, and work full time labeling and evaluating data. To decrease inter-rater discrepancy and ensure that our instructions were clear, we had an initial calibration session where they test-rated a small portion of the data (around 10 datapoints which were then omitted from the results) and asked us any clarifying questions. For each style, we compare outputs from our method plus the three baselines for 50 sentences.

Each tuple was scored by three raters (3,600 ratings total) on the following three axes which are standard to textual style transfer \citep{DBLP:journals/corr/abs-1904-02295}: \textbf{(1) transfer strength} (the amount that the output actually matches the target style), \textbf{(2) semantic preservation} (whether the underlying meaning of the output text, aside from style, matches that of the input), and \textbf{(3) fluency} (whether the text is coherent and could have been written by a proficient English speaker). Following \citet{sakaguchi-van-durme-2018-efficient}, transfer strength and semantic preservation were rated on a scale from 1--100. A screenshot of the evaluation UI is shown in Figure \ref{fig:rater_ui} in the Appendix. Note that the guidelines for semantic preservation are not standardized in prior literature \citep{DBLP:journals/corr/abs-2106-04747}; while some evaluations are strict that the outputs cannot contain any more information than the inputs, we asked the annotators not to penalize for meaning transformations which are necessary for the specified transformation.
We use \textit{dialog-LLM}, and compare it with three other methods: \textbf{(1) zero-shot} (a baseline), \textbf{(2) paraphrase} (our normal augmented zero shot prompt, but with the target style of \textit{``paraphrased''}, as a control) and \textbf{(3) human} (ground-truth transformations written by the authors).

\begin{figure}
    \includegraphics[width=\linewidth]{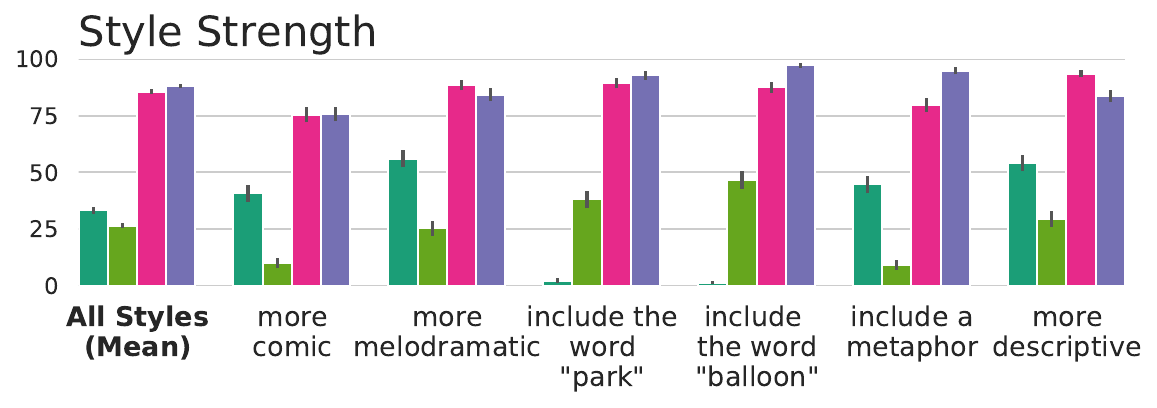}
    \includegraphics[width=\linewidth]{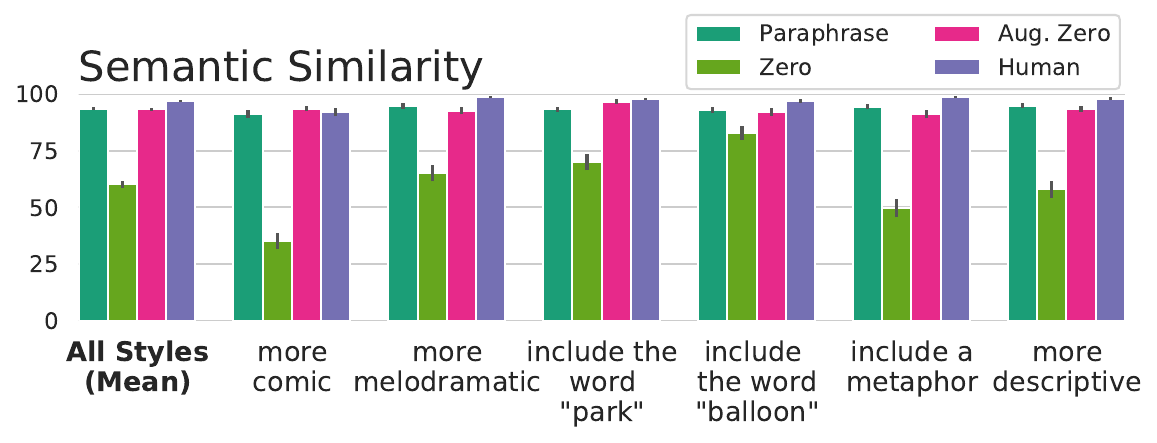}
  \vspace{-0.7cm}
  \caption{Human evaluation of style transfer for six atypical styles. Our method is rated comparably to the human-written ground truth. Error bars show Standard Error of the Mean. Evaluation of fluency is shown in Figure \ref{human_eval_other_styles_fluency} in the Appendix.}
  \vspace{-0.1cm}
  \label{human_eval_other_styles}
\end{figure}

Figure \ref{human_eval_other_styles} shows these results. 
We found that the outputs of our method were rated almost as highly as the human-written ground truth for all three evaluations. The zero-shot baseline performed the worst in all categories: 25.4\% of the time, it did not return a valid response at all (see $\S$\ref{section:limitations}), compared with 0.6\% for augmented zero shot.
The strong performance of the paraphrase baseline at fluency and semantic similarity shows that large LMs are capable of generating high quality text that remains true to the input sentence's meaning.
Overall, the average length of the input sentences was 66 characters, whereas the average length of augmented zero-shot outputs was 107 characters. For context, human paraphrase outputs were 82 characters.

For a subset of the tasks, some automatic evaluation was also possible.
We found that the ``\textit{balloon}'' and ``\textit{park}'' transformations successfully inserted the target word 85\% of the time.
For ``\textit{more descriptive}'' and ``\textit{include a metaphor}'' the transformed text was, as expected, longer than the original (by 252\% and 146\% respectively, compared with 165\% and 146\% for human baselines).

\subsection{Standard Styles}
\begin{figure}[t]
    \includegraphics[width=\linewidth]{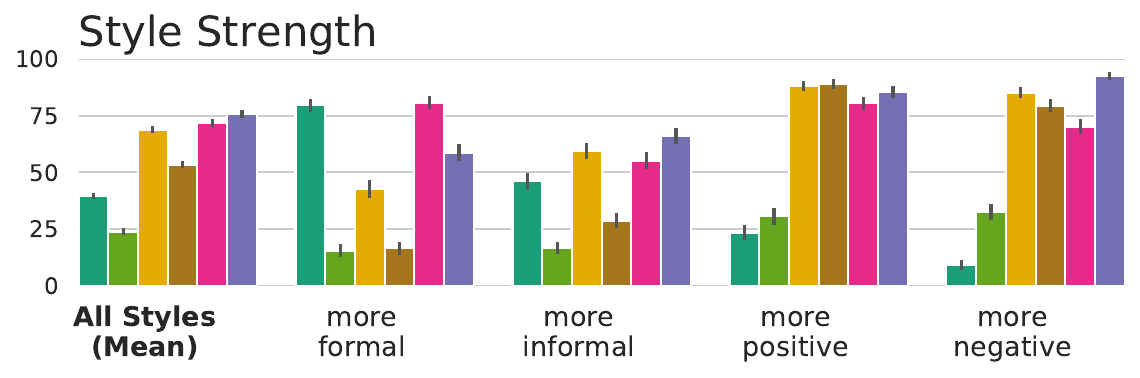}
    \includegraphics[width=\linewidth]{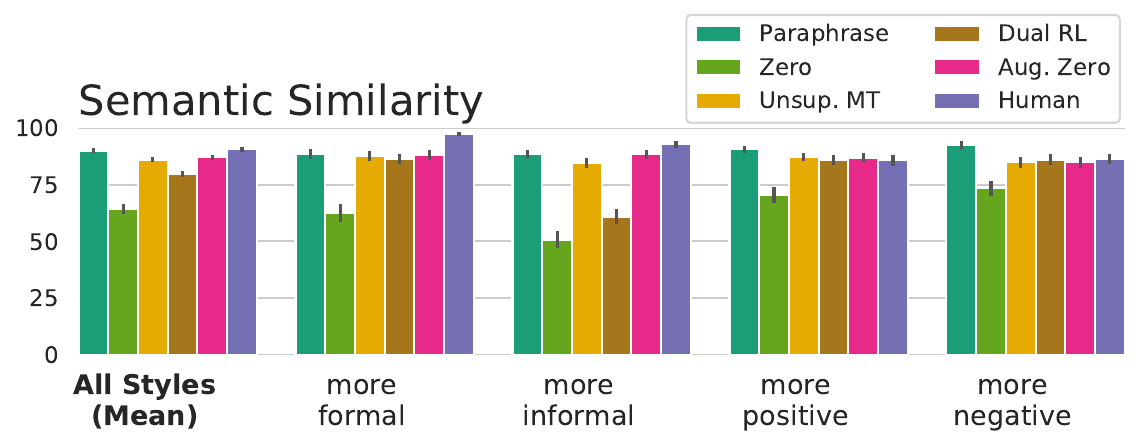}
  \caption{Human evaluation of sentiment and formality transfer. Our method is rated comparably to human-written ground truth as well as prior methods. Error bars show Standard Error of the Mean. Unsup.~MT is \citet{{prabhumoye-etal-2018-style}}; Dual RL is \citet{DBLP:conf/ijcai/LuoLZYCSS19}. }
  \label{human_eval_standard_style}
\end{figure}

To better contextualize the performance of our method with prior methods, we also generated outputs for two standard style transfer tasks: sentiment and formality. 
Figure \ref{human_eval_standard_style} shows human evaluations (same setup as before) for our outputs as well as the outputs from two popular prior style transfer methods, Unsup MT \citep{prabhumoye-etal-2018-style} and Dual RL \citep{DBLP:conf/ijcai/LuoLZYCSS19}.
The outputs from our method were rated comparably to both human generated responses and the two prior methods, using the same rating setup as the non-standard styles, with six outputs and baselines for four styles across 50 sentences, rated independently by three raters, totalling 3,000 total ratings.

Furthermore, following \citet{li-etal-2018-delete} and \citet{sudhakar-etal-2019-transforming}, we perform automatic evaluation for sentiment style transfer since there are classifiers available for these styles. We note that although automatic evaluations can diverge from human ratings, they can still be a good proxy as we could not perform human evaluation against every prior method due to time and resource constraints.
We automatically evaluate \textbf{(1) transfer strength} using a sentiment classifier from HuggingFace Transformers \citep{wolf-etal-2020-transformers}, \textbf{(2) semantic similarity} to human examples provided by \citet{DBLP:conf/ijcai/LuoLZYCSS19} via BLEU score, and \textbf{(3) fluency} via perplexity, as measured by GPT-2 (117M).

\begin{table}[t]
    \setlength{\tabcolsep}{3pt}
    \centering \small 
        \begin{tabular}{l r r r}
        \toprule
         & Acc & BLEU & PPL\\
        \midrule
        \underline{\textsc{Supervised Methods}} \\
        Cross-alignment \citep{NIPS2017_2d2c8394} & 73.4 & 17.6 & 812 \\
        Backtrans \citep{prabhumoye-etal-2018-style} & 90.5 & 5.1 & 424 \\
        Multidecoder \citep{AAAI1817015}  & 50.3 & 27.7 & 1,703\\
        Delete-only \citep{li-etal-2018-delete} & 81.4 & 28.6 & 606 \\
        Delete-retrieve \citep{li-etal-2018-delete} & 86.2 & 31.1 & 948\\
        Unpaired RL \citep{xu-etal-2018-unpaired} & 52.2 & 37.2 & 2,750\\
        Dual RL \citep{DBLP:conf/ijcai/LuoLZYCSS19} & 85.9 & 55.1 & 982\\
        Style transformer \citep{dai-etal-2019-style} & 82.1 & 55.2 & 935\\
        \midrule
        \underline{\textsc{Inference-only methods}} \\
        GPT-3 ada, aug zero-shot& 31.5 & 39.0 & 283\\
        GPT-3 curie, aug zero-shot& 53.0 & 48.3 & 207\\
        GPT-3 da vinci, aug zero-shot& 74.1 & 43.8 & 231\\
        LLM: zero-shot & 69.7 & 28.6 & 397 \\ 
        {\color{white}LLM: }five-shot & 83.2 & 19.8 & 240 \\
        {\color{white}LLM: }aug zero-shot & 79.6 & 16.1 & 173 \\
        LLM-dialog: zero-shot & 59.1 & 17.6 & 138 \\
        {\color{white}LLM-dialog: }five-shot & 94.3 & 13.6 & 126 \\
        {\color{white}LLM-dialog: }aug zero-shot & 90.6 & 10.4 & 79 \\
        \bottomrule
        \end{tabular}
\caption{
Comparing augmented zero-shot prompting with supervised style transfer methods on the Yelp sentiment style transfer dataset using automatic evaluation.
Acc: accuracy; PPL: perplexity.  The inference-only table shows our method applied to 3 different sizes of GPT-3, plus our own LLM. 
}
        \label{tab:summary_table}
\end{table}

Table \ref{tab:summary_table} shows these automatic evaluations, with four main takeaways. 
First, augmented zero-shot prompting achieves high accuracy and low perplexity compared with baselines. The BLEU scores, however, are low, which we believe is because it tends to add additional information to generated sentences (see Appendix \ref{appendix:low_bleu} for a deeper analysis).
Second, we apply augmented zero-shot learning to GPT-3 175B; these results indicate that augmented zero-shot learning generalizes to another large language model.
Third, we vary model size for GPT-3 models, finding that larger size greatly improves style transfer. 
Fourth, for \textit{LLM} and \textit{LLM-dialog}, we find that augmented zero-shot learning substantially outperforms vanilla zero-shot learning and almost reaches the accuracy of five-shot learning. 

\section{Potential of Arbitrary Styles}\label{subsec:potential}
 
\begin{table}[t]
\small
\def\arraystretch{1.5}%
\begin{tabular}{p{1.0\columnwidth}}
\hline
to be a little less angsty • to be about mining • to be better written • to be less diabolical • to be more absurd • to be more adventurous • to be more Dickensian • to be more emotional • to be more magical • to be more melodramatic • to be more philosophical • to be more revolutionary • to be more surprising • to be more suspenseful • to be more technical • to be more whimsical • to be warmer • to fit better grammatically with the rest of the story • to make more sense \\
\hline
\end{tabular}
\caption{\label{tab:unique_requests}Requests in the form of ``\textit{Rewrite this...}'' made by real users to a large LM-powered text editor. For the full set of unique requests, see Table \ref{tab:unique_requests_full} in the Appendix.}
\label{tab:realusers}
\end{table}

One promising application of augmented zero-shot learning is an AI-powered writing assistant that can allow writers to transform their text in arbitrary ways that the writer defines and controls.
As a qualitative case study to explore what arbitrary re-write styles may be requested, we built an AI-assisted story-writing editor with a ``rewrite as'' feature that uses our augmented few-shot method.
Our editor has a freeform text box for users to specify how they would like a selection of their story to be rewritten (see Figure \ref{fig:wc} in the Appendix).
We asked 30 people from a creative writing group to use our UI to write a 100-300 word story,  collecting 333 rewrite requests in total. 
Table \ref{tab:unique_requests} shows a subset of these, which were as diverse as asking for the text ``\textit{to be about mining}'' or ``\textit{to be less diabolical}.''

\section{Limitations and Failure Modes}
\label{section:limitations}
This section details several qualitative limitations with our method.
\paragraph{Unparsable answers} A frequent problem that arises when using large LMs for other NLP tasks is their outputs cannot be automatically parsed into usable answers. For example, when given a prompt like \begin{small}
 \texttt{``Here is some text: {that is an ugly dress}. Here is a rewrite of the text, which is more positive''} \end{small}
 \textit{LLM-Dialog} might return something like \begin{small}
 \texttt{``Sounds like you are a great writer!''} \end{small} Similar error modes exist for \textit{LLM}, which might output something like \begin{small}
 \texttt{``Here are more writing tips and tricks.''} \end{small} Other times, the response contains correct information, but it cannot be automatically parsed (e.g., \begin{small}
 \texttt{``a good rewrite might be to say that the dress is pretty.''} \end{small}) In hindsight, these outputs make a lot of sense: most of the training data of large LMs is not well-formatted pairs of inputs and outputs  \citep{reynolds2021prompt}. See $\S$\ref{sec:prompt-selection} for how we dealt with these issues.

\paragraph{Hallucinations} Large LMs are known to hallucinate text content; we saw this happen frequently for style transfer. While this is an advantage in some contexts like creative writing, it is undesirable for applications like summarization. 
\paragraph{Inherent style trends} We also noticed that even our \textit{``paraphrase''} baseline, where the model was simply asked to rewrite the input sentence, was rated highly for style strength for a few styles, including \textit{``more formal''} and \textit{``more melodramatic''}.
This implies that our method's generations generally trend toward these styles.
A direction for future work would be to see what styles and qualities of text our method (and large LMs in general) are inherently more likely to produce.

\paragraph{Less reliable than trained methods}
For style transfer tasks that have available training data, prior methods that either train or finetune on that data are going to be inherently more reliable at producing text that looks like their training data.
This can be observed in the lower BLEU scores our method achieves than trained methods, despite comparable transfer accuracy (Section \ref{appendix:low_bleu}).
Thus, augmented zero-shot learning offers less fine-grained controllability in the properties of the style-transferred text than methods which see task-specific training data.

\paragraph{Large LM safety concerns} Large LMs themselves come with their own host of difficulties, barriers to entry, and potential safety concerns as discussed by \citet{10.1145/3442188.3445922}, which are also valid for this style transfer method. However, we also think that this method can be a useful tool in exploring and exposing the safety and boundaries of these models themselves: what happens if we try to force the large LM to make a text ``more racist'', ``more sexist'', or ``more incendiary''? It is important to keep pushing these models to their boundaries to see where they fail and where problems arise, and specific use cases that show a broader range of the model's capabilities also show a broader range of its failure modes.

\section{Conclusions}
We introduced augmented zero-shot learning, which we find shows shows strikingly promising performance considering its simplicity.
This prompting paradigm moves the needle in text style transfer by expanding the range of possible styles beyond the currently limited set of styles for which annotated data exists.
More broadly, we also hope that the strategy of prompting a large LM with non-task specific examples can inspire new inference-only methods for other NLP tasks.

\bibliography{references}
\bibliographystyle{acl_natbib}

\clearpage
\appendix

\section*{Appendix}
\label{sec:appendix}

\section{Prompt Selection}
\label{sec:prompt-selection}
A promising new area of prompt engineering has arisen to address the failure modes discussed above, specifically the invalid or unparseable answers. \citet{reynolds2021prompt} find that prompting a model for a task is more akin to locating an already-learned task than truly learning a new one. Moreover, they emphasize that prompt engineering is mostly about avoiding various failure cases such as those described above.
 In this work, we use delimiters (``\{'' and ``\}'') to help avoid these types of errors, giving scores of zero when there was no valid responses with such delimiters. There are other delimiters that could be used (e.g., quotes, ``('' and ``)'', ``<'' and ``>'', newlines with a colon (as used by GPT-3), etc. We chose curly braces as they were 1) likely to occur in the training data as delimiters in other contexts and 2) not frequently part of the input sentence itself. We also use a second person prompt template for the dialog, which yielded better results as it was more similar to the training data. Exploring these options more quantitatively would be an interesting direction for future work.
Because the performance of prompting can vary depending on the exact language of the prompt \cite{reynolds2021prompt}, we compare four variations of prompts for sentiment: ``\textit{more {positive/negative}},'' ``\textit{{happier/sadder}},''
``\textit{more {optimistic/pessimistic}},'' and
``\textit{more {cheerful/miserable}}.''
As shown in Table~\ref{tab:compare-between-prompts} in the Appendix, performance differed across the four prompts, but we found them comparable.
\begin{table}[ht]
    \centering \small 
    \begin{tabular}{lrrr}
    \toprule
    Model / prompt wording &Acc&Bleu&PPL \\
    \midrule
    \underline{LLM}&&&\\
    ``more {positive/negative}''&76.3&14.8&180\\
    ``{happier/sadder}''&62.6&15.5&173\\
    ``more {optimistic/pessimistic}''&69.7&14.1&143\\
    ``more {cheerful/miserable}''&74.5&15.7&186\\
    \midrule
    \underline{LLM-Dialog}&&&\\
    ``more {positive/negative}''&90.5&10.4&79 \\
    ``{happier/sadder}''&85.9&9.6&90 \\
    ``more {optimistic/pessimistic}''&85.8&10.2&79 \\
    ``more {cheerful/miserable}''&88.8&11.4&93 \\
    \bottomrule
    \end{tabular}
\caption{Comparing variations of augmented zero-shot learning prompt wording for sentiment style transfer.}
\label{tab:compare-between-prompts}
\end{table}

\begin{table}[t]
\small
\def\arraystretch{1.5}%
\begin{tabular}{p{0.9\columnwidth}}
\hline
into paragraphs • to be a bit clearer • to be a little less angsty • to be a word for a song • to be about mining • to be about vegetables • to be better written • to be less descriptive • to be less diabolical • to be more absurd • to be more adventurous • to be more angry • to be more cheerful • to be more descriptive • to be more Dickensian • to be more emotional • to be more fancy • to be more flowery • to be more interesting • to be more joyful • to be more magical • to be more melodramatic • to be more philosophical • to be more revolutionary • to be more scary • to be more subtle • to be more surprising • to be more suspenseful • to be more technical • to be more violent • to be more whimsical • to be warmer • to fit better grammatically with the rest of the story • to make more sense • to use a more interesting word • with a few words \\
\hline
\end{tabular}
\caption{ Full results for requests in the form of ``\textit{Rewrite this...}'' made by users to a large LM-powered text editor. }
\label{tab:unique_requests_full}
\end{table}

\begin{table*}[ht]
    \small
    \centering
    \def\arraystretch{1.5}%
    \begin{tabular}{p{0.41\linewidth} p{0.45\linewidth}}
    \toprule
      \textbf{Input} & \textbf{Model suggestion} \\ \hline
      You honor me,'' said Bruno. As he spoke, he began to laugh. His laugh was merry sounding and melodic. He took a deep breath. When he spoke again, his voice was low and strong. ``I take this great task on and with my brothers shall lead us on to victory! \newline \textbf{Rewrite it to be about mining.} & You honor me, said Bruno. As he spoke, he began to laugh. His laugh was deep and gravely, as if from the bowels of the earth. He took a deep breath. When he spoke again, his voice was deep and dark.`I take this great task on and with my brothers shall delve into the mines!'' \\ \hline
      I reckon,' smiled Toe, 'you'd have to say the same for icicles, wouldn't you? \newline \textbf{Rewrite it to be more interesting.} & ‘I reckon,’ smiled Toe, ‘you’d have to say the same for icicles, wouldn’t you?’ These words were barely out of his mouth before the others realized exactly what they meant.\\ \hline
      with its southern perfume \newline \textbf{Rewrite it to be more descriptive} & A mixture of sweet honeysuckles and jasmine  with  a  hint  of  tobacco  and warm mesquite smoke. It was his southern perfume, with a hint of a memory of simpler times.\\ \hline
      But Sally remained unmoved--and lonely. \newline \textbf{Rewrite it to be more emotional} & Her tears rolled and rolled down her cheek, mixing with the saltwater around her, until the water was a rich mixture of the two. She wanted her poem to be cheerful like it used to be, but couldn't find the right words. \\
    \bottomrule
    \end{tabular}
    \caption{Examples of users' arbitrary style transfer requests for which the model suggestion was accepted.}
    \label{tab:selected_usage_examples}
    \vspace*{3mm}
\end{table*}

\begin{table*}[ht]
    \centering
    \tiny
    \begin{tabular}{p{.9\linewidth}}
\hline
\multicolumn{1}{c}{Augmented Zero-shot Prompt: LLM}
\\
\texttt{Here is some text: \{When the doctor asked Linda to take the medicine, he smiled and gave her a lollipop.\}. Here is a rewrite of the text, which is more scary. \{When the doctor told Linda to take the medicine, there had been a malicious gleam in her eye that Linda didn't like at all.\} Here is some text: \{they asked loudly, over the sound of the train.\}. Here is a rewrite of the text, which is more intense. \{they yelled aggressively, over the clanging of the train.\} Here is some text: \{When Mohammed left the theatre, it was already dark out\}. Here is a rewrite of the text, which is more about the movie itself. \{The movie was longer than Mohammed had expected, and despite the excellent ratings he was a bit disappointed when he left the theatre.\} Here is some text: \{next to the path\}. Here is a rewrite of the text, which is about France. \{next to la Siene\} Here is some text: \{The man stood outside the grocery store, ringing the bell.\}. Here is a rewrite of the text, which is about clowns. \{The man stood outside the circus, holding a bunch of balloons.\} Here is some text: \{the bell ringing\}. Here is a rewrite of the text, which is more flowery. \{the peales of the jangling bell\} Here is some text: \{against the tree\}. Here is a rewrite of the text, which is include the word "snow". \{against the snow-covered bark of the tree\}} 
\textcolor{blue}{\texttt{Here is some text: \{That is an ugly dress\}. Here is a rewrite of the text, which is more positive."}}
\\
\hline
\multicolumn{1}{c}{Augmented Zero-shot Prompt: LLM-dialog}
\\
\texttt{> Here is some text: \{When the doctor asked Linda to take the medicine, he smiled and gave her a lollipop.\}. Rewrite it to be more scary. \newline
> \{When the doctor told Linda to take the medicine, there had been a malicious gleam in her eye that Linda didn't like at all.\} \newline
> Here is some text: \{they asked loudly, over the sound of the train.\}. Rewrite it to be more intense. \newline
> \{they yelled aggressively, over the clanging of the train.\} \newline
> Here is some text: \{When Mohammed left the theatre, it was already dark out\}. Rewrite it to be more about the movie itself. \newline
> \{The movie was longer than Mohammed had expected, and despite the excellent ratings he was a bit disappointed when he left the theatre.\} \newline
> Here is some text: \{next to the path\}. Rewrite it to be about France. \newline
> \{next to la Siene\} \newline
> Here is some text: \{The man stood outside the grocery store, ringing the bell.\}. Rewrite it to be about clowns. \newline
> \{The man stood outside the circus, holding a bunch of balloons.\} \newline
> Here is some text: \{the bell ringing\}. Rewrite it to be more flowery. \newline
> \{the peals of the jangling bell\} \newline
> Here is some text: \{against the tree\}. Rewrite it to be include the word "snow". \newline
> \{against the snow-covered bark of the tree\} \newline
}
\textcolor{blue}{\texttt{> Here is some text: \{That is an ugly dress\}. Rewrite it to be more positive."}}
\\
\hline
\end{tabular}
    \caption{In black, we show the exact augmented-zero shot prompts used in our experiments, for \textit{LLM} and GPT-3 (top), and for \textit{LLM-Dialog} (bottom). As shown, for \textit{LLM-Dialog}, we replaced ``\texttt{Here is a rewrite of the text, which is}'' with ``\texttt{Rewrite it to be}''. Each line starting with ``>"" above was passed in as an individual dialog turn. 
    The blue shows how an input text and goal style are concatenated to the few-shot prompt in order to produce final model output.
    Note that we can achieve high accuracy even though the prompt formulation resulted in some minor grammatical errors for some styles (e.g., ``\texttt{rewrite it to be include the word 'snow'}''). Text versions of these prompts can be downloaded at \url{https://bit.ly/3fLDuci}.}
    \label{tab:fullprompt}
\end{table*}

\section{Low BLEU for \textit{LLM} Outputs}\label{appendix:low_bleu}
As we saw in Table \ref{tab:summary_table}, the outputs of our model had low BLEU scores with respect to human generated outputs, while simultaneously having high semantic similarity in human evaluations. 
Based on qualitative examination of outputs, we believe that this is because model outputs often, despite having high semantic similarity with the source sentence, used different language from human annotations.
For instance, for transferring the sentiment of ``\textit{ever since joes has changed hands it's just gotten worse and worse}'' to positive sentiment, our augmented zero-shot learning model outputted ``{the establishment has continued to provide excellent service, improving steadily since its change of ownership}.'' 
This will have low BLEU with the ground truth with respect to human references, which is simply ``\textit{ever since joes has changed hands it's just gotten better and better}.'' 

Though we do not see this as an inherent problem, increasing the BLEU for the purposes of comparison can be done in an easy way via candidate selection, as our model returns sixteen possible continuations. In applications for which we prefer model outputs to have high lexical similarity to the source sentence, we could select the candidate of the sixteen with the highest BLEU score compared with the original source sentence. 
We find that this candidate selection step can substantially improve the BLEU score with the ground truth target sentences, as we show in Table \ref{tab:candidate_select}. 

\begin{table}
    \centering \small 
        \begin{tabular}{l r r r}
        \toprule
         & Acc & BLEU & PPL\\
        \midrule
        \underline{LLM-128B} \\
        Zero-shot & 69.7 & 28.6 & 397 \\
        + cand.~select. & 31.4 & 61.5 & 354\\
        Five-shot & 83.2 & 19.8 & 240\\
        + cand.~select. & 61.5 & 55.6 & 306\\
        Augmented zero-shot & 79.6 & 16.1 & 173\\
        + cand.~select. & 65.0 & 49.3 & 292\\
        \midrule
        \underline{LLM-128B-dialog} \\
        Zero-shot & 59.1 & 17.6 & 138\\
        + cand.~select. & 46.8 & 24.2 & 166\\
        Five-shot & 94.3 & 13.6 & 126\\
        + cand.~select. & 81.3 & 47.6 & 345\\
        Augmented zero-shot & 90.6 & 10.4 & 79\\
        + cand.~select. & 73.7 & 40.6 & 184\\
        \bottomrule
        \end{tabular}
\caption{
Sentiment style transfer results with candidate selection (cand.~select.). 
Candidate selection means that of the sixteen examples returned by our model, we choose the one with the highest BLEU with the source sentence.  
}
\label{tab:candidate_select}
\end{table}

\section{Further Related Work}
Style transfer has gained increasing attention in the NLP landscape, for which neural models have been trained to perform style transfer for styles including sentiment, formality, politeness, gender, and political slant \citep{prabhumoye-etal-2018-style,madaan-etal-2020-politeness,10.1145/3449139}.
We will briefly summarize the primary approaches to style transfer here, and refer the involved reader to either \cite{DBLP:journals/corr/abs-2011-00416} or \cite{DBLP:journals/corr/abs-2010-12742} for a survey.

Most text style transfer approaches fall in two categories. 
Early approaches tend to require \textit{parallel} text data \citep{zhu-etal-2010-monolingual,rao-tetreault-2018-dear}, where every input in the source style has a corresponding output in the target style.
Though this formulation elegantly fits the standard encoder--decoder paradigm, the availability of a parallel text corpus is a stringent requirement. 
Hence, recent text style transfer approaches have instead used \textit{non-parallel} monostyle data (no one-to-one-mapping between instances in the source and target styles). 
Such methods include latent representation manipulation \citep{liu2020revision}, prototype-based text editing \citep{li-etal-2018-delete}, and pseudo-parallel corpus construction \citep{jin-etal-2019-imat}.
However, even non-parallel monostyle data can be hard to collect for arbitrary styles. 
As such, surveys have called for more research on approaches that expand the scope of supported styles and reduce the training data requirements for style transfer systems 
\citep{DBLP:journals/corr/abs-2011-00416,DBLP:journals/corr/abs-2010-12742}.

Several new methods tackle the challenging problem of \textit{label-free} style transfer, which does not require a full corpus of labeled data, but rather just a few exemplars that define a style. 
\citet{DBLP:journals/corr/abs-1905-11975} use variational autoencoders for unsupervised learning of controllable representations for text.
\citet{DBLP:journals/corr/abs-2010-03802} extract a style vector from a set of target texts and use this vector to condition the decoder to perform style transfer to a target style. 
These approaches have a similar goal to ours in terms of expanding the scope of possible style transfers. However, they are different in two main ways. First, they require a fully specialized model, where our method can be applied out-of-the-box with something like GPT-3. This can either be a strength or weakness, depending on the availability of such a model. Second, they require exemplars to define a style rather than a plain text description.

\begin{figure}
    \includegraphics[width=\linewidth]{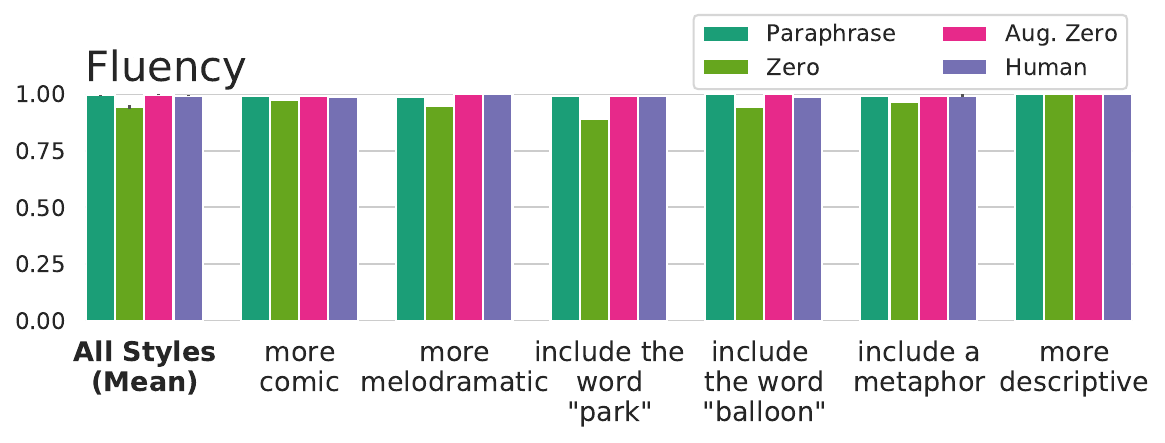}
    \includegraphics[width=\linewidth]{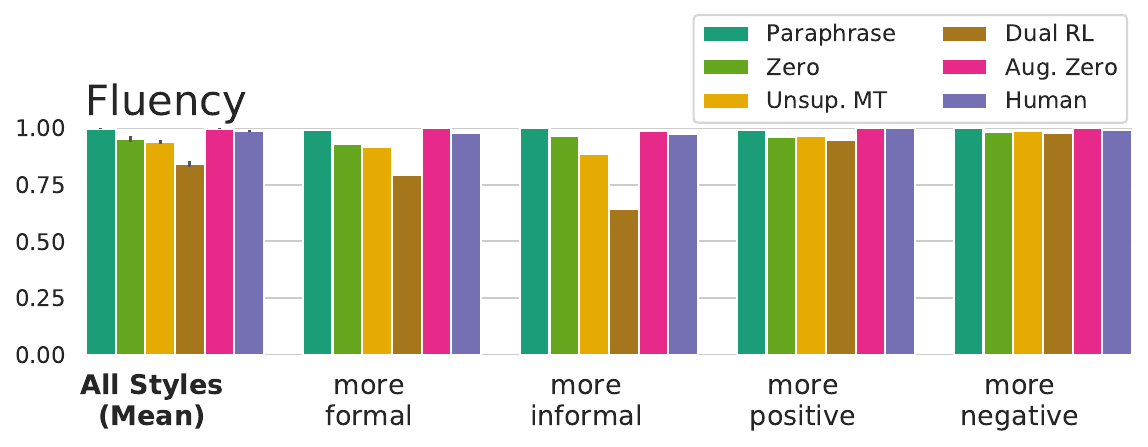}
  \caption{Human evaluation of fluency for style transfer for six atypical styles. Error bars show standard error of the mean.}
  \vspace{-0.1cm}
  \label{human_eval_other_styles_fluency}
\end{figure}

\begin{figure*}[!htb]
  \centering
  \fbox{\includegraphics[width=1\linewidth]{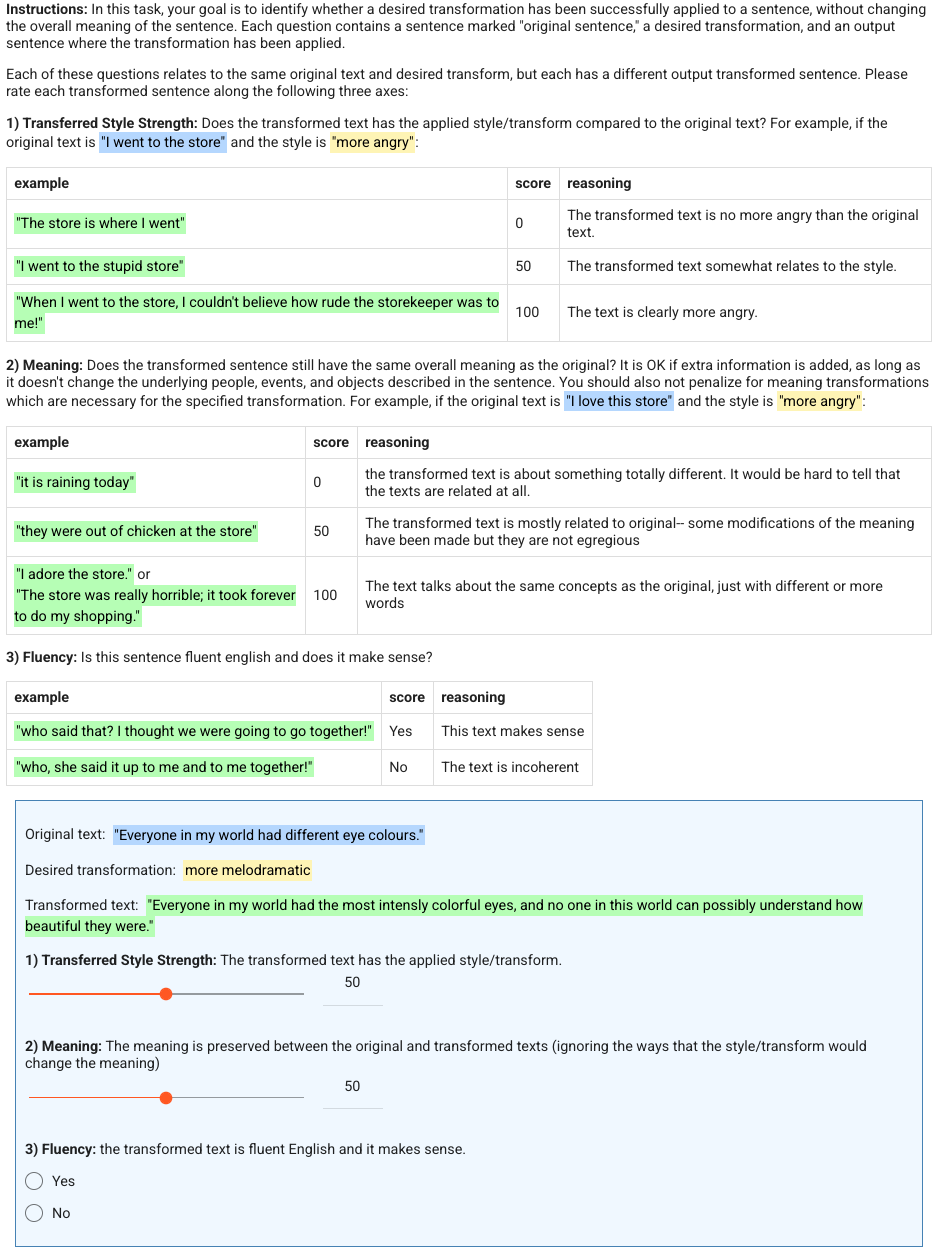}}
  \caption{The rating UI used for human evaluation. The user may be shown a number of blue squares at once with the same original text and different outputs.}
  \label{fig:rater_ui}
  \vspace{-0.1cm}
\end{figure*} 

\begin{figure*}[!htb]
  \centering
  \fbox{\includegraphics[width=1\linewidth]{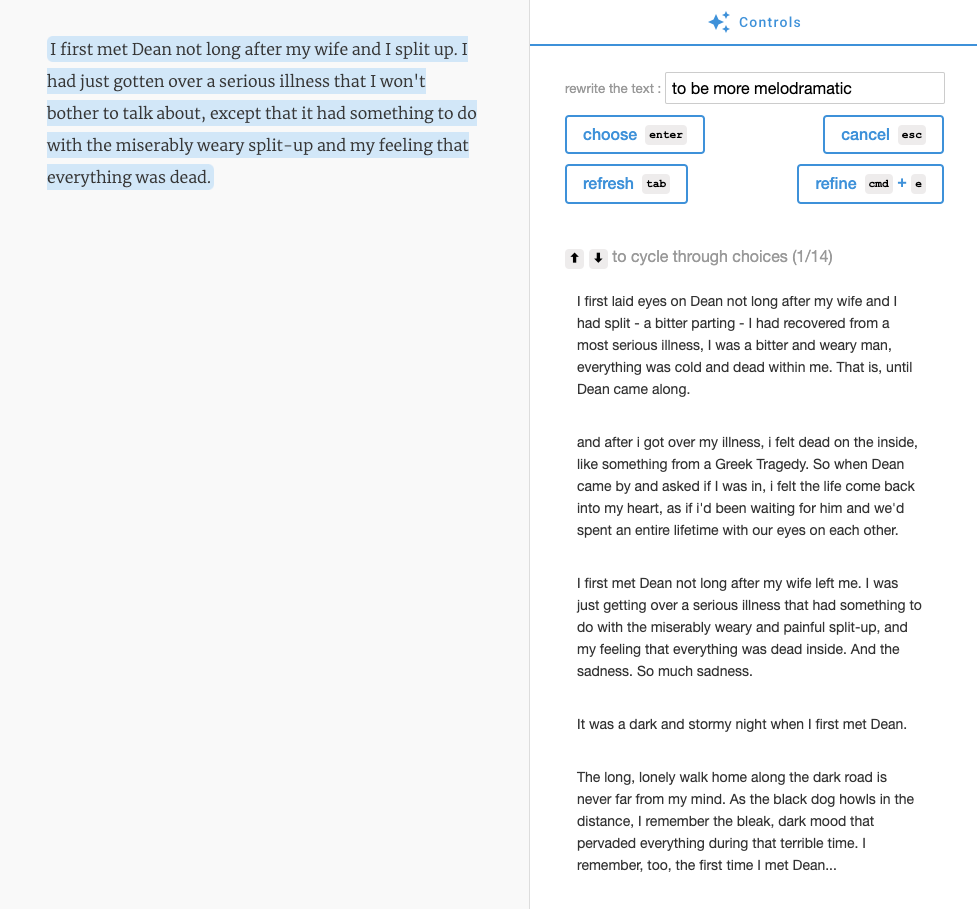}}
  \caption{Screenshot AI-assisted editor with `Rewrite as' feature.\label{fig:wc}}
  \vspace{-0.1cm}
\end{figure*} 

\begin{table*}[]
    \centering
    \small
    \begin{tabular}{l|r|rrrr}
    \toprule
    Style & Inputs & Aug. Zero & Zero & Human & Paraphrase \\
    \hline
        more comic &75&116&63&97&87\\
        more melodromatic &75&124&88&116&87 \\
        include the word ``park'' &75&124&72&94&87\\
        include the word ``balloon'' &75&135&86&98&87\\
        include a metaphor &75&110&74&110&87 \\
        more descriptive &75&190&105&124&87\\
        \hline
        Overall &75&133&81&107&87 \\
        \bottomrule
    \end{tabular}
    \caption{The mean length in characters of the inputs and outputs for our six atypical styles.}
    \label{tab:lengths}
\end{table*}

\end{document}